\documentclass[11pt,a4paper]{article}

\usepackage[utf8]{inputenc} 
\usepackage{amsmath, amssymb, amsthm} 
\usepackage{bm} 
\usepackage{graphicx} 
\usepackage{caption} 
\usepackage{titlesec} 
\usepackage[top=2.5cm, bottom=2.5cm, left=3cm, right=3cm]{geometry} 
\usepackage{fancyhdr} 
\usepackage{booktabs} 
\usepackage{array} 
\usepackage{float}

\newcommand{\norm}[1]{\left\lVert #1 \right\lVert}

\newcommand{\tr}[1]{}

\newcommand{\Rb}{\mathbb{R}}

\usepackage{stmaryrd}

\titleformat{\section}{\large\bfseries}{\thesection}{1em}{}
\titleformat{\subsection}{\normalsize\bfseries}{\thesubsection}{1em}{}

\DeclareMathOperator{\diag}{diag}

\title{Error-State LQR Formulation for Quadrotor UAV Trajectory Tracking}
\author{Micah Reich \\ Carnegie Mellon University \\ \texttt{mreich@cmu.edu}}
\date{\today}

\begin{document}

\maketitle

\begin{abstract}
    This article presents an error-state Linear Quadratic Regulator (LQR) formulation for robust trajectory tracking in quadrotor Unmanned Aerial Vehicles (UAVs). The proposed approach leverages error-state dynamics and employs exponential coordinates to represent orientation errors, enabling a linearized system representation for real-time control. The control strategy integrates an LQR-based full-state feedback controller for trajectory tracking, combined with a cascaded bodyrate controller to handle actuator dynamics. Detailed derivations of the error-state dynamics, the linearization process, and the controller design are provided, highlighting the applicability of the method for precise and stable quadrotor control in dynamic environments.
\end{abstract}

\section{Introduction}

\begin{flushleft}
    The control of quadrotor Unmanned Aerial Vehicles (UAVs) presents unique challenges due to their nonlinear dynamics, underactuation, and the need for precise trajectory tracking in dynamic environments. Traditional control techniques often struggle to handle these challenges efficiently while maintaining computational tractability for real-time applications. To address these issues, this work outlines an error-state Linear Quadratic Regulator (LQR) approach, leveraging the compact and singularity-free representation of orientation errors using exponential coordinates.
\end{flushleft}

\begin{flushleft}
    Exponential coordinates provide a robust way to represent orientation errors without the singularities inherent in other parameterizations such as Euler angles. By formulating the controller in terms of error-state dynamics, this approach avoids the complexity of directly controlling the nonlinear dynamics, focusing instead on minimizing deviations from a nominal trajectory. This is achieved by driving the error-state—which includes position, velocity, and orientation errors—toward zero.    
\end{flushleft}

\begin{flushleft}
    The proposed controller uses an LQR formulation, a well-established concept in classical control theory for linear systems, to minimize a quadratic cost function balancing state deviations and control effort. Although the quadrotor dynamics are nonlinear, the error-state dynamics can be re-linearized about the current tracking error at a sufficiently high frequency, allowing the LQR controller to operate effectively in real time. This iterative re-linearization ensures that the controller remains responsive to changes in the tracking error while maintaining computational efficiency.
\end{flushleft}

\newpage
\section{Definitions and system dynamics}

\subsection{State and control definition table} \label{sec:defn_table}

\begin{flushleft}
    The state of the quadrotor UAV is defined as follows, inspired by \cite{8798359,solà2017quaternionkinematicserrorstatekalman}:
    
    \begin{center}
        \setlength{\arrayrulewidth}{0.2mm} 
        \renewcommand{\arraystretch}{1.1}  
        \setlength{\tabcolsep}{8pt}        

        \begin{tabular}{|c|c|c|c|c|}
            \hline
            Name & True & Nominal & Error & Composition \\
            \hline
            Full state & $\bm{x}_t$ & $\bm{x}$ & $\delta \bm{x}$ & $\bm{x}_t = \bm{x} \oplus \delta\bm{x}$ \\
            Position & $\bm{p}_t$ & $\bm{p}$ & $\delta \bm{p}$ & $\bm{p}_t = \bm{p} + \delta \bm{p}$ \\
            Velocity & $\bm{v}_t$ & $\bm{v}$ & $\delta \bm{v}$ & $\bm{v}_t = \bm{v} + \delta \bm{v}$ \\
            Orientation & $\bm{q}_t$ & $\bm{q}$ & $\delta \bm{q}$ & $\bm{q}_t = \bm{q} \otimes \delta \bm{q}$ \\
            Rotation Matrix & $\bm{R}_t$ & $\bm{R}$ & $\delta \bm{R}$ & $\bm{R}_t = \bm{R} \,\delta \bm{R}$\\
            Exponential Vector & & & $\delta \bm{\theta}$ & $\delta \bm{R} = \exp\left([\delta \bm{\theta}]_{\times}\right)$ \\
            \hline
        \end{tabular}
    \end{center}

    The full state of the quadrotor UAV includes the position and velocity of the center of mass, expressed in the inertial frame, as well as the orientation represented as a unit quaternion:
    \[
    \bm{x} = \begin{pmatrix} \bm{p} & \bm{q} & \bm{v} \end{pmatrix}^T \in \Rb^3 \times S^3 \times \Rb^3.
    \]
    The error-state contains the $\delta$-values for these quantities, with the important distinction that the orientation error component is represented in exponential coordinates:
    \[
    \delta \bm{\theta} \in \Rb^3, \quad \delta \bm{x} = \begin{pmatrix} \delta \bm{p} & \delta \bm{\theta} & \delta \bm{v} \end{pmatrix}^T \in \Rb^9.
    \]
\end{flushleft}

\begin{flushleft}
    The nominal state of the quadrotor UAV may be derived from trajectory optimization processes or the differential flatness property of quadrotor dynamics \cite{5980409}.
\end{flushleft}

\begin{flushleft}
    The control of the quadrotor UAV is defined as follows:
    
    \begin{center}
        \setlength{\arrayrulewidth}{0.2mm} 
        \renewcommand{\arraystretch}{1.1}  
        \setlength{\tabcolsep}{8pt}        

        \begin{tabular}{|c|c|c|c|c|}
            \hline
            Magnitude & True & Nominal & Error & Composition \\
            \hline
            Full control & $\bm{u}_t$ & $\bm{u}$ & $\delta \bm{u}$ & $\bm{u}_t = \bm{u} \oplus \delta\bm{u}$ \\
            Collective Thrust & $c_t$ & $c$ & $\delta c$ & $c_t = c + \delta c$ \\
            Angular Velocity & $\bm{\omega}_t$ & $\bm{\omega}$ & $\delta \bm{\omega}$ & $\bm{\omega}_t = (\delta \bm{R})^T\bm{\omega} + \delta \bm{\omega}$ \\
            \hline
        \end{tabular}
    \end{center}

    The full control of the quadrotor UAV contains the collective thrust and desired angular velocity expressed in the
    current body frame, $\bm{u} = \begin{pmatrix} c & \bm{\omega}\end{pmatrix}^T \in \Rb^4$. The error-state contains the
    $\delta$-values for these quantities, taking care to ensure that the angular velocity error is represented in the current
    body frame (with the nominal angular velocity represented in the nominal frame's coordinates), with
    $\delta \bm{u} = \begin{pmatrix} \delta c & \delta \bm{\omega}\end{pmatrix}^T \in \Rb^4$.
\end{flushleft}

\subsection{The true-state kinematics and dynamics}
The true-state kinematics and dynamics follow from the single rigid-body assumption and lumped-mass model of the quadrotor UAV, with thrust
acting along the body $z$-axis and gravity acting along the world $-z$ axis.

\begin{equation}
    \dot{\bm{p}}_t = \bm{v}_t
\end{equation}
\begin{equation}
    \dot{\bm{v}}_t = \bm{g} + \frac1{m} \bm{R}_t \begin{pmatrix}
        0 \\ 0 \\ c_t
    \end{pmatrix}
\end{equation}
\begin{equation}
    \dot{\bm{q}}_t = \frac12 \bm{q}_t \otimes \begin{pmatrix}
        0 \\ \bm{\omega}_t
    \end{pmatrix}
\end{equation}

\subsection{The nominal-state kinematics and dynamics}
\begin{flushleft}
    The nominal-state kinematics and dynamics are identical to those of the true state, with appropriate substitutions:
    \begin{equation}
        \dot{\bm{p}} = \bm{v}
    \end{equation}
    \begin{equation}
        \dot{\bm{v}} = \bm{g} + \frac1{m} \bm{R} \begin{pmatrix}
            0 \\ 0 \\ c
        \end{pmatrix}
    \end{equation}
    \begin{equation}
        \dot{\bm{q}} = \frac12 \bm{q} \otimes \begin{pmatrix}
            0 \\ \bm{\omega}
        \end{pmatrix}
    \end{equation}
\end{flushleft}

\subsection{The error-state kinematics and dynamics}

\begin{flushleft}
    We now derive the time derivatives of the error-state $\dot{\delta \bm{x}}$ for each component. Throughout this derivation, we assume the error-state is small, ignore second-order terms, and employ small-angle approximations. First, for the position error:
    \begin{align*}
        \bm{p}_t &= \bm{p} + \delta \bm{p}, \\
        \delta \bm{p} &= \bm{p}_t - \bm{p}, \\
        \dot{\delta \bm{p}} &= \dot{\bm{p}_t} - \dot{\bm{p}}, \\
        &= \bm{v}_t - \bm{v}, \\
        &= \delta \bm{v}.
    \end{align*}
\end{flushleft}

\begin{flushleft}
    We next solve for $\dot{\delta \bm{v}}$, using the approximation $\bm{R}_t = \bm{R} (I + [\delta \bm{\theta}]_\times) + O(\norm{\delta \bm{\theta}}^2)$:
    \begin{align*}
        \bm{v}_t &= \bm{v} + \delta \bm{v} \\
        \delta \bm{v} &= \bm{v}_t - \bm{v} \\
        \dot{\delta \bm{v}} &= \dot{\bm{v}_t} - \dot{\bm{v}} \\
        &= \bm{g} + \frac1{m} \bm{R}_t \begin{pmatrix}
            0 \\ 0 \\ c_t
        \end{pmatrix} -
        \left(
            \bm{g} + \frac1{m} \bm{R} \begin{pmatrix}
                0 \\ 0 \\ c
            \end{pmatrix}
        \right) \\
        &= \frac1m \bm{R}_t \begin{pmatrix}
            0 \\ 0 \\ c_t
        \end{pmatrix}
        -
        \frac1m \bm{R} \begin{pmatrix}
            0 \\ 0 \\ c
        \end{pmatrix} \\
        &= \frac1m \left(
            \bm{R} \, \delta \bm{R} \begin{pmatrix}
                0 \\ 0 \\ c_t
            \end{pmatrix}
            -
            \bm{R} \begin{pmatrix}
                0 \\ 0 \\ c
            \end{pmatrix}
        \right) \\
        &\approx
        \frac1m \left(
            \bm{R} (I + [\delta \bm{\theta}]_\times) \begin{pmatrix}
                0 \\ 0 \\ c_t
            \end{pmatrix}
            -
            \bm{R} \begin{pmatrix}
                0 \\ 0 \\ c
            \end{pmatrix}
        \right) \tag{Small angle approximation of $\delta \bm{R}$}  \\
        &= \frac1m \left(
            (\bm{R} + \bm{R}[\delta \bm{\theta}]_\times)\begin{pmatrix}
                0 \\ 0 \\ c_t
            \end{pmatrix}
            -
            \bm{R} \begin{pmatrix}
                0 \\ 0 \\ c
            \end{pmatrix}
        \right) \\
        &= \frac1m \left(
            \bm{R} \begin{pmatrix}
                0 \\ 0 \\ c_t
            \end{pmatrix} -
            \bm{R} \begin{pmatrix}
                0 \\ 0 \\ c
            \end{pmatrix} + 
            \bm{R}[\delta \bm{\theta}]_\times \begin{pmatrix}
                0 \\ 0 \\ c_t
            \end{pmatrix}
        \right) \\
        &= \frac1m \bm{R} \left(
            \begin{pmatrix}
                0 \\ 0 \\ \delta c
            \end{pmatrix}
            +
            [\delta \bm{\theta}]_\times
            \begin{pmatrix}
                0 \\ 0 \\ c + \delta c
            \end{pmatrix}
        \right)
    \end{align*}
\end{flushleft}

\begin{flushleft}
    Lastly, we solve for $\dot{\delta \bm{\theta}}$.
    We note that the angular velocity of a rigid body expressed in the body frame and the exponential coordinates 
    for the body frame orientation are related by:
    \begin{align*}
        \dot{\bm{\theta}} &= J_r^{-1}(\bm{\theta}) \bm{\omega}
        \quad
        \mathrm{with}
        \quad
        J_r^{-1}(\bm{\theta}) =
        I + \frac12 [\bm{\theta}]_\times + \left(
            \frac{1}{\norm{\bm{\theta}}^2}
            -
            \frac{1 + \cos \norm{\bm{\theta}}}
            {2\norm{\bm{\theta}}\sin \norm{\bm{\theta}}}
        \right)
        [\bm{\theta}]_\times^2
    \end{align*}
    where $J_r^{-1}(\bm{\theta})$ is the inverse of the right Jacobian of $SO(3)$ \cite{solà2021microlietheorystate}. Using this fact, we may write:
    \begin{align*}
        \dot{\delta \bm{\theta}} &= J_r^{-1}(\delta \bm{\theta}) \delta \bm{\omega} \\
        &\approx (I + \frac12 [\delta \bm{\theta}]_\times) \delta \bm{\omega}
        \tag{Small angle approximation of $J_r^{-1}(\delta \bm{\theta})$}
    \end{align*}
\end{flushleft}

\section{Linearization of the error-state dynamics}

\subsection{Approximate linearization derivation}

\begin{flushleft}
    For a nonlinear function $f : \Rb^{n_x} \times \Rb^{n_u} \to \Rb^{n_x}$, the first-order Taylor approximation
    of $f(\bm{x}, \bm{u})$ about some point $(\bm{\bar{x}}, \bm{\bar{u}})$ is written as:

    \begin{align*}
        f(\bm{x}, \bm{u}) &\approx 
        f(\bm{\bar{x}}, \bm{\bar{u}}) +
        \frac{\partial f}{\partial \bm{x}} \Bigg|_{\bar{\bm{x}}, \bar{\bm{u}}} (\bm{x} - \bm{\bar{x}}) +
        \frac{\partial f}{\partial \bm{u}} \Bigg|_{\bar{\bm{x}}, \bar{\bm{u}}} (\bm{u} - \bm{\bar{u}}) \\
        &=
        \left(
            f(\bm{\bar{x}}, \bm{\bar{u}}) -
            \frac{\partial f}{\partial \bm{x}} \Bigg|_{\bar{\bm{x}}, \bar{\bm{u}}} \bm{\bar{x}} -
            \frac{\partial f}{\partial \bm{u}} \Bigg|_{\bar{\bm{x}}, \bar{\bm{u}}} \bm{\bar{u}}
        \right)
        +
        \frac{\partial f}{\partial \bm{x}} \Bigg|_{\bar{\bm{x}}, \bar{\bm{u}}} \bm{x} + 
        \frac{\partial f}{\partial \bm{u}} \Bigg|_{\bar{\bm{x}}, \bar{\bm{u}}} \bm{u}
    \end{align*}

    Notice that the expression:
    \[
        f(\bm{\bar{x}}, \bm{\bar{u}}) -
        \frac{\partial f}{\partial \bm{x}} \Bigg|_{\bar{\bm{x}}, \bar{\bm{u}}} \bm{\bar{x}} -
        \frac{\partial f}{\partial \bm{u}} \Bigg|_{\bar{\bm{x}}, \bar{\bm{u}}} \bm{\bar{u}}
    \]
    is nothing but the first-order Taylor approximation of $f$ about $(\bm{\bar{x}}, \bm{\bar{u}})$ evaluated at 
    the point $(\bm{0}, \bm{0})$, which we can take to be near-zero for $f$ with $f(\bm{0}, \bm{0}) = \bm{0}$ and a linearization point 
    $(\bm{\bar{x}}, \bm{\bar{u}})$ sufficiently close to 0.
\end{flushleft}

\begin{flushleft}
    Thus, for $f(\bm{0}, \bm{0}) = \bm{0}$ and $(\bm{\bar{x}}, \bm{\bar{u}})$ sufficiently close to 0, we may approximate $f(\bm{x}, \bm{u})$ as:
    \[
        f(\bm{x}, \bm{u}) \approx 
        A \bm{x} +
        B \bm{u}
        \quad
        \textrm{where}
        \quad
        A \in \Rb^{n_x \times n_x},
        \,\,
        B \in \Rb^{n_x \times n_u}
    \]
    with:
    \[
        A = \frac{\partial f}{\partial \bm{x}} \Bigg|_{\bar{\bm{x}}, \bar{\bm{u}}}
        \quad
        \textrm{and}
        \quad
        B = \frac{\partial f}{\partial \bm{u}} \Bigg|_{\bar{\bm{x}}, \bar{\bm{u}}} 
    \]
\end{flushleft}

\begin{flushleft}
    For our purposes, we will write $\dot{\delta \bm{x}} = f(\delta \bm{x}, \delta \bm{u})$ and write the linearized system as:
    \[
        \dot{\delta \bm{x}} \approx
        \frac{\partial f}{\partial \delta \bm{x}} \Bigg|_{\overline{\delta \bm{x}}, \overline{\delta \bm{u}}}
        \delta \bm{x}
        +
        \frac{\partial f}{\partial \delta \bm{u}} \Bigg|_{\overline{\delta \bm{x}}, \overline{\delta \bm{u}}}
        \delta \bm{u}
    \]
    where $\overline{\delta \bm{x}}$ is the current error state and $\overline{\delta \bm{u}}$ is the current error control.
\end{flushleft}

\subsection{Jacobians of error-state dynamics}

\begin{flushleft}
    We write out the relevant Jacobians to construct the $A$ and $B$ matrices of the linearized error-state dynamics.
    These matrices take the form:

    \[
        \frac{\partial f}{\partial \delta \bm{x}} =
        \begin{pmatrix}
            0 & 0 & \partial \dot{\delta \bm{p}} / \partial \delta \bm{v} \\
            0 & \partial \dot{\delta \bm{\theta}} / \partial \delta \bm{\theta} & 0 \\
            0 & \partial \dot{\delta \bm{v}} / \partial \delta \bm{\theta} & 0
        \end{pmatrix}
        \quad
        \frac{\partial f}{\partial \delta \bm{u}} =
        \begin{pmatrix}
            0 & 0 \\
            0 & \partial \dot{\delta \bm{\theta}} / \partial \delta \bm{\omega} \\
            \partial \dot{\delta \bm{v}} / \partial \delta c & 0
        \end{pmatrix}
    \]

    From the above expressions for the error-state dynamics, we take derivatives and arrive at:
    \begin{align*}
        \frac{\partial \dot{\delta \bm{p}}}{\partial \delta \bm{v}} &= I \in \Rb^{3 \times 3} \\
        \frac{\partial \dot{\delta \bm{\theta}}}{\partial \delta \bm{\theta}} &= -\frac12 [\delta \bm{\omega}]_\times \in \Rb^{3 \times 3} \\
        \frac{\partial \dot{\delta \bm{v}}}{\partial \delta \bm{\theta}} &= -
        \frac1m \bm{R} \left[
            \begin{pmatrix}
                0 \\ 0 \\ c + \delta c
            \end{pmatrix}
        \right]_\times \in \Rb^{3 \times 3} \\
        \frac{\partial \dot{\delta \bm{\theta}}}{\partial \delta \bm{\omega}} &= I + \frac12 [\delta \bm{\theta}]_\times \in \Rb^{3 \times 3} \\
        \frac{\partial \dot{\delta \bm{v}}}{\partial \delta c} &= 
        \frac1m \bm{R} 
        \left(
            I
            + 
            [\delta \bm{\theta}]_\times
        \right)
        \begin{pmatrix}
            0 \\ 0 \\ 1
        \end{pmatrix}
        \in \Rb^{3 \times 1}
    \end{align*}
\end{flushleft}

\section{Error-state LQR controller}

\subsection{LQR controller formulation}
\begin{flushleft}
    We design an LQR controller to drive the error state to zero, where these quantities are
    defined in section~\ref{sec:defn_table}.
    The LQR full-state feedback gain matrix $\bm{K}$ is obtained by minimizing the
    inifinite-horizon cost function $J(\delta \bm{x}, \delta \bm{u})$:
    \[
        J(\delta \bm{x}, \delta \bm{u}) =
        \int_0^{\infty} (
            \delta \bm{x}^T Q \delta \bm{x} +
            \delta \bm{u}^T R \delta \bm{u}
        ) dt
    \]
    for $Q, R \succ 0$. The cost function is minimized by the control policy $\delta \bm{u} = -\bm{K} \delta \bm{x}$. 
    where $\bm{K}$ is given by:
    \begin{align*}
        \bm{K} &= R^{-1} B^T P
    \end{align*}
    with $P$ being the solution to the continuous-time Algebraic Riccati Equation.
    The full control at time $t$ is then given by:
    \[
        \bm{u}_t = \bm{u} - \bm{K} \delta \bm{x}
    \]
    As a practical implementation consideration, one must ensure that the $A$ matrix of the
    linearized error-state system is stable, i.e. $\Re(\lambda_i) < 0$ for all eigenvalues $\lambda_i$
    in the spectrum of $A$. A small amount of regularization may be added to $A$ to ensure the ARE has a solution.
\end{flushleft}

\begin{figure}[H]
    \centering
    \includegraphics[width=1.0\textwidth]{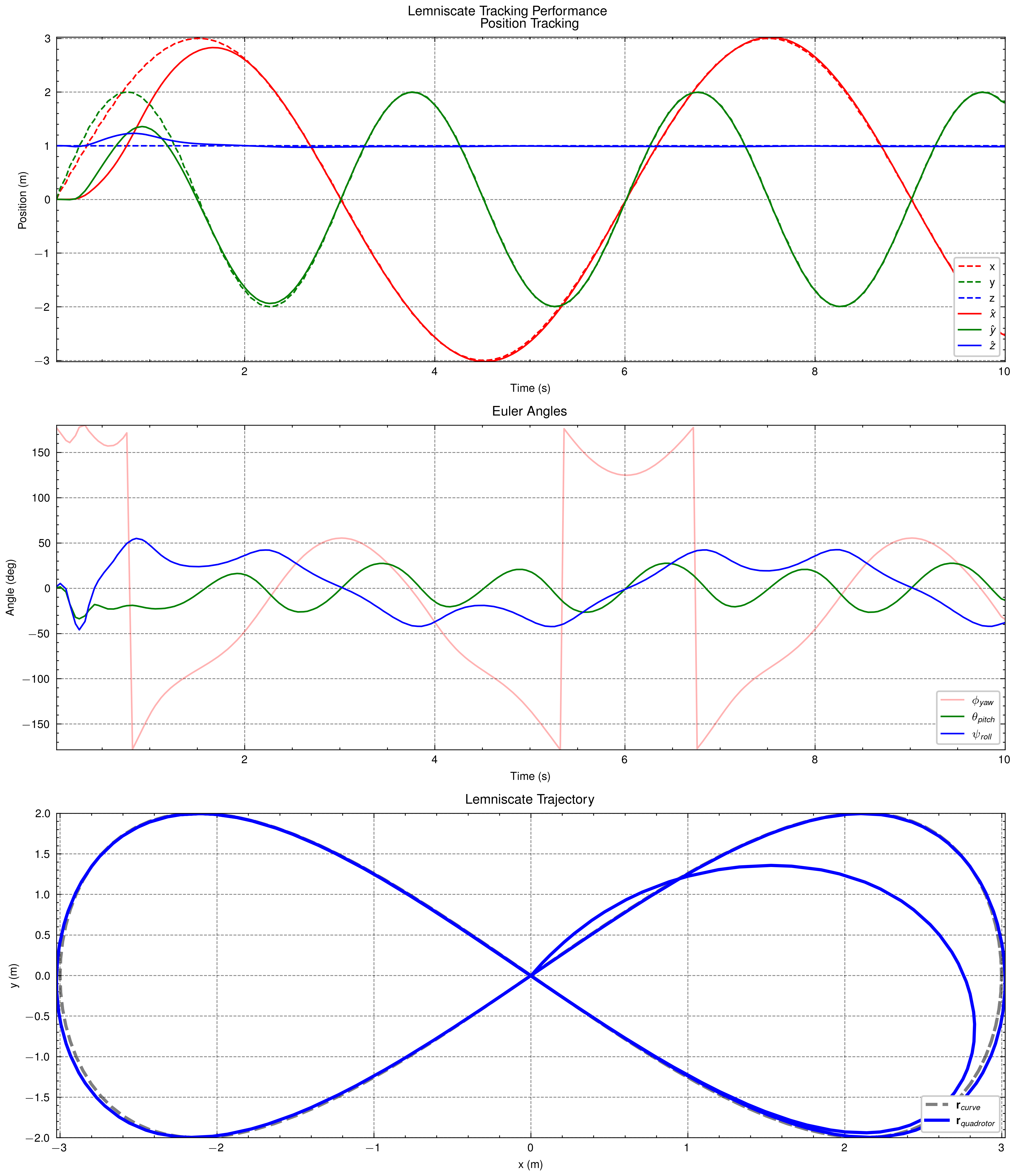}
    \caption{Tracking performance of the error-state LQR controller on a lemniscate trajectory
    while tracking yaw angles. The tracking error at the beginning of the trajectory is due to
    the fact that the UAV begins with a flat initial orientation $q_0 = (1, 0, 0, 0)^T$.}
    \label{fig:control_diagram}
\end{figure}

\subsection{Bodyrate controller formulation}

\begin{flushleft}
    Using the principle of time-scale separation, the LQR controller uses
    the desired angular velocity as a virtual control input to the system, similar to \cite{8460885}.
    In reality, the control inputs are the collective thrust and the body torque.\footnote{Often, the inputs are considered to be the four motor speeds, but we will
    consider the collective thrust and body torque for simplicity. The mapping between
    motor speeds and collective wrench is usually written as $\bm{F} = G \bm{\Omega}^2$
    where $\bm{\Omega}^2$ is the vector of squared motor angular velocities.}
\end{flushleft}

\begin{flushleft}
    In a cascaded fashion, we use a simple bodyrate P-controller with feedback linearization to track the desired angular velocities produced
    by the LQR controller. The bodyrate controller is given by:
    \[
        \bm{\tau} = J (K_p (\bm{\omega} - \bm{\omega}_t)) +
        \bm{\omega}_t \times J \bm{\omega}_t
    \]
    where $J$ is the inertia tensor of the quadrotor UAV and $K_p = \diag(K_{p,1}, K_{p,2}, K_{p,3})$ is the proportional gain matrix.
\end{flushleft}

\bibliographystyle{plain}
\bibliography{references}

\end{document}